\documentclass[lettersize,journal]{IEEEtran}
\usepackage{amsmath,amsfonts}
\usepackage{algorithmic}
\usepackage{algorithm}
\usepackage{array}
\usepackage[caption=false,font=normalsize,labelfont=sf,textfont=sf]{subfig}
\usepackage{textcomp}
\usepackage{stfloats}
\usepackage{url}
\usepackage{verbatim}
\usepackage{graphicx}
\usepackage{cite}

\usepackage{xcolor}
\usepackage{verbatim} 

\usepackage{xspace}

\makeatletter
\DeclareRobustCommand\onedot{\futurelet\@let@token\@onedot}
\def\@onedot{\ifx\@let@token.\else.\null\fi\xspace}

\def\eg{\emph{e.g}\onedot} 
\def\ie{\emph{i.e}\onedot}

\makeatother

\begin{document}

\title{A Brain-Inspired \\ Perception-Decision Driving Model \\Based on Neural Pathway Anatomical Alignment}

\author{Haidong Wang, Pengfei Xiao, Ao Liu, Qia Shan, Jianhua Zhang
}

\markboth{Journal of \LaTeX\ Class Files,~Vol.~14, No.~8, August~2021}%
{Shell \MakeLowercase{\textit{et al.}}: A Sample Article Using IEEEtran.cls for IEEE Journals}


\maketitle

\begin{abstract}
In the realm of autonomous driving, conventional approaches for vehicle perception and decision-making primarily rely on sensor input and rule-based algorithms. 
However, these methodologies often suffer from lack of interpretability and robustness, particularly in intricate traffic scenarios. 
To tackle this challenge, we propose a novel brain-inspired driving (BID) framework. 
Diverging from traditional methods, our approach harnesses brain-inspired perception technology to achieve more efficient and robust environmental perception. 
Additionally, it employs brain-inspired decision-making techniques to facilitate intelligent decision-making. 
The experimental results show that the performance has been significantly improved across various autonomous driving tasks and achieved the end-to-end autopilot successfully. 
This contribution not only advances interpretability and robustness but also offers fancy insights and methodologies for further advancing autonomous driving technology.
\end{abstract}
    
\section{Introduction}
\label{sec:intro}
\hspace{1pc}Autonomous driving \cite{prakash2020exploring,li2024brain} is an advanced technology that intelligent vehicles perceive road environments through onboard sensor systems, autonomously plan driving routes, and control vehicles to reach predetermined destinations. 
Its technical system generally includes three major parts: environmental perception, decision planning, and vehicle control~\cite{Zhang:2021}, involving multiple research fields such as computer science, mathematics, mechanical engineering, control science, and psychology~\cite{Codevilla:2019}.

However, the current autonomous driving systems still suffer from insufficient interpretability due to the existence of ``black box" nature of deep learning models \cite{ohn2020learning}, greatly limiting the credibility and widespread application of various perception and decision-making methods in practical engineering. 
Even though the use of generative adversarial networks~\cite{zugner2020adversarial} to generate explanatory data related to decision-making has been attempted, the quality of such data is often substandard, and the training process is quite challenging. 
Furthermore, modern urban traffic conditions are characterized by high dynamics and strong uncertainty. 
When the environment is complex, the autonomous driving system may experience performance degradation. 
Therefore, to overcome the limitations of existing methods, there is a need to design more interpretable and robust autonomous driving systems, thereby ensuring the safety of vehicle operation.
\begin{figure}[t]
	\centering
	\includegraphics[width=0.95\linewidth]{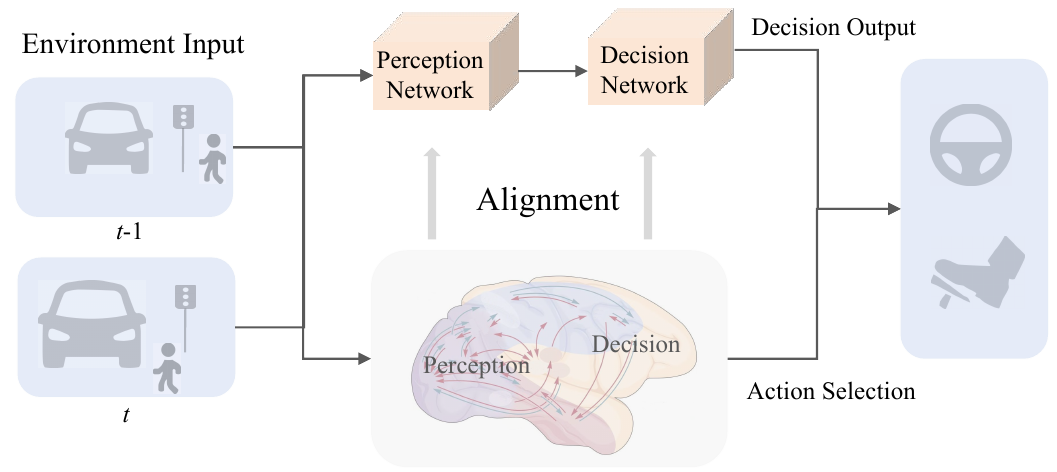}
	\caption{Perception-decision network diagram based on neural pathway anatomical alignment}
	\label{fig:fig1}
\end{figure}

In this work, we propose a novel brain-inspired driving (BID) framework inspired by brain perception and decision-making. 
Unlike previous methods, our proposed BID expert does not rely on manually formulated rules and demonstrates strong interpretability and robustness.
Our work involves constructing a brain-inspired perception network for extracting environmental features and a brain-inspired decision-making network for generating driving decisions for the target vehicle, in order to train the BID expert~\cite{kahn2021land}. 
Its structure is as shown in Fig.~\ref{fig:fig1}. 
Specifically, the brain-inspired perception network consists of a visual recognition~\cite{8574054}, a motion perception, and a multimodal fusion~\cite{liao2021statistical}. 
The brain-inspired decision-making network comprises a decision generation~\cite{dai2013dynamic} and a decision evaluation. 
Driving decisions encompass multiple driving actions of the target vehicle, and the BID agent directly imitates expert behavior. 
Overall, our main contribution lies in proposing a novel brain-inspired perception model and a brain-inspired decision-making model for autonomous driving, aiming to achieve a more robust and interpretable autonomous driving system.

\section{Related Work}

\subsection{Brain-Inspired Perception}
\hspace{1pc}The human visual cortex possesses remarkable environmental perception capabilities, serving as a crucial component of the central nervous system and responsible for transforming visual signals into comprehensible information \cite{9134376}. 
Traditional visual encoding models are limited by the use of either handcrafted or deep learning features merely, despite their individual advantages\cite{kubilius2019brain}. 
To integrate the two, Cui et al. \cite{8574054} proposed the GaborNet-VE model, which combines Gabor features with deep learning to form an efficient visual encoding framework. 
However, this model requires significant parameter adjustments across different tasks, and the interpretability of its deep learning component remains inadequate\cite{liao2021statistical}. 
In contrast, our proposed BID model demonstrates outstanding performance in revealing decision-making mechanisms and exhibiting strong generalization capabilities.

\subsection{Brain-Inspired Decision}
\hspace{1pc}In practical environments, agents often need to handle continuous state spaces, while traditional reinforcement learning models, such as Q learning methods, are more fit to discrete states\cite{xi2020automatic}. 
To address this issue, Zhao et al. \cite{zhao2018brain} proposed the prefrontal cortex and basal ganglia method, which subdivides continuous states and utilizes a continuous function to capture temporal reward. 
However, the design of continuous reward functions remains a challenge. 
In non-Gaussian and nonlinear environments, traditional methods struggle to cope with their nonlinear and non-Gaussian characteristics\cite{naghshvarianjahromi2020natural}. 
Inspired by human decision-making in brain, some researchers~\cite{dai2013dynamic} designed an autonomous computation layer accroding to a cognitive dynamic system (CDS), providing agents in NGNLE with stronger decision-making capabilities. 
Although these methods attempt to mimic the decision-making process of the human brain, structural differences lead to a lack of transparency in their processes. 
Therefore, it is crucial to construct a network architecture that align with the anatomical structure of human brain.

\section{Method}

\subsection{Task Formulation}

\hspace{1pc}We use imitation learning (IL) to train the proposed BID.
In IL, expert first demonstrates a driving action $a_{t}$ (such as steering, brake and throttle) in response to an input frame $f_t$ based on the expert's policy $\pi(f_{t})$.
It reflects the expert's driving skills, judgement, and behavior.
The core idea of IL is to train the proposed BID agent to replicate the expert's actions by learning from these observed interactions.

\begin{figure*}[t]
	\centering
	\includegraphics[width=\linewidth]{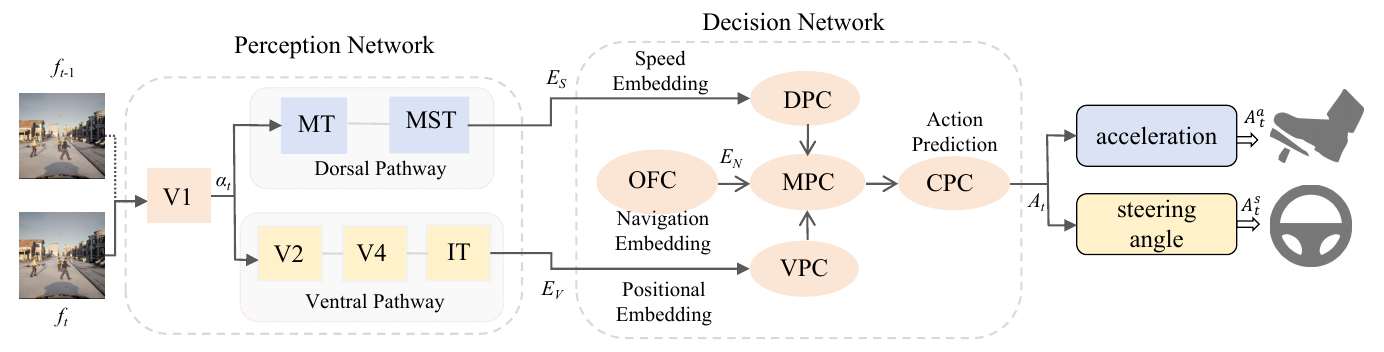}
	\caption{The network architecture of the BID agent is primarily composed of the perception network and the decision network.
		The perception module includes both the dorsal and ventral pathways, which allow the system to understand and process its surroundings.
		The decision network consists of five modules that represent different regions of the prefrontal cortex: the medial prefrontal cortex (MPC), orbital prefrontal cortex (OPC), caudal prefrontal cortex (CPC), dorsal prefrontal cortex (DPC), and ventral prefrontal cortex (VPC). 
		Specifically, the input to the system is an RGB image $f_{t} $, which represents the current visual scene.
		The output is the action $A_t$, which controls the ego-vehicle's maneuvering directly.
		The action $A_t$ includes the steering angle $ A_t^s $ and acceleration $ A_t^a $ of the vehicle, making BID a brain-inspired and end to end autonomous driving model.
		Besides, in the construction process of training dataset, the driving actions and input frames are paired.
	}
	\label{fig:fig2}
\end{figure*}

Before training the BID using IL, we first collect a series of frame and action pairs $(f_{t}, a_{t})$ produced by expert demonstrations.
These data are then utilized to train the BID policy $\pi_{\theta}(f_{t})$, which approximates the expert's policy.
%
In the validating phase, we no longer utilize the expert's strategy, but only the trained BID policy.

\subsection{Neural Aligned BID}
\hspace{1pc}Inspired by the brain's remarkable capabilities in perception and decision-making, we have proposed the perception and decision network based on neural pathway anatomical alignment to simulate the information processing mechanisms of brain in an autonomous driving system. 
This approach is designed to significantly improve both the interpretability and robustness of the system.

\subsubsection{Perception Network}
\hspace{1pc}As shown in Fig.~\ref{fig:fig2}, our brain-inspired perception network begins by capturing precise input data from the surrounding environment, which includes essential traffic elements such as roads, vehicles, pedestrians, and traffic signals. 
At time $t$, the input frame $f_t$ is processed through a series of deep convolutional neural networks and recurrent networks, incorporating activation functions commonly used in deep neural networks. 
This processing mirrors the way external information is handled by the brain's visual cortex\cite{kubilius2019brain}. 
Initially, the primary visual cortex (V1) processes the visual input, extracting key features. 
To manage the high computational demands, the first area, V1$_\text{BID}$, includes a 7$\times$7 convolution operation and another 3$\times$3 convolution. 
These features are then passed along to the dorsal visual pathway, where middle temporal (MT) and middle superior temporal (MST) specialize in encoding velocity information between current frame $ f_t $ and previous frame $ f_{t-1} $, particularly related to spatial location and movement\cite{wang2022btn}. 
The dorsal pathway is implemented as a dynamic filter network\cite{jia2016dynamic} and utilized to extract the spatial relation between background and foreground, 
and uses appearance repsentation $ \alpha_t $ to dynamically infer convolutional filter.

At the same time, the ventral pathway receives these features, with inferior temporal (IT) cortex focusing on encoding object category information, thereby facilitating object recognition.
Each visual area in ventral pathway is simulated with a specific module, where neurons perform some classical, standard operations, such as convolution, nonlinearity, batch normalization, and max pooling on its input. 
The structure of the pathway is same across all visual areas, though the number of cells for a submodule may vary.
The subsequent areas, V2$_\text{BID}$, V4$_\text{BID}$ and IT$_\text{BID}$, each carry out two 1$\times$1 convolutional opeartion, a 3 $ \times $ 3 bottleneck style convolutional operation, and another 1$\times$1 convolution. 
To simulate recurrent connection, the output for each submodule are fed back into the same submodule 4 times for further processing.

After this sequence of processing steps, our network is capable of efficiently extracting and outputting multi-dimensional embeddings of the current environment to next decision network. 
These features include both the appearance characteristics $ E_V $ of objects and motion-related information $ E_S $.

\subsubsection{Decision Network}
The output features from the perception network are then efficiently passed to a brain-inspired decision-making network, where they undergo further processing through transformer networks, along with anatomical alignment to the prefrontal cortex pathway. 
This simulates how sensory information is processed by different regions of the prefrontal cortex in the brain.
%

The ventral prefrontal cortex (VPC) generates goals based on recognition embedding (visual cues), allowing for quick responses and adaptation to changes in the environment\cite{passingham2012neurobiology}.
In our setup, we flatten the embedding patches and then fed it to transformer models.
We use the classical one-dimensional position feature to offer positional context, and add this embedding directly to the token.

The dorsal prefrontal cortex (DPC) is responsible for generating goals based on sequences, as well as temporal and spatial environmental cues, offering clear guidance for decision-making and action\cite{passingham2012neurobiology}.
DPC is implemented with a dynamic filter network~\cite{jia2016dynamic} and is used to address spatial relations between the foreground and background.
Within this process, the orbitofrontal cortex (OFC) plays a crucial role in establishing future action goals $n_t$, which is encoded as a one-hot command embedding.
It helps to clarify and focus on the desired objectives. 
The navigation command $n_t$ is mapped into a vector embedding $E_N$ using a fully connected layer (FC).

The medial prefrontal cortex (MPC) is critical for making decisions based on current states, along with a high-level navigation command\cite{passingham2012neurobiology}.
It ensure that our actions are optimized according to previous experiences and future goal.
The input embeddings of MPC are merged with the output of VPC, DPC and OFC.
We format the BID with the input data $O_t=(f_t, n_t)$, parameterized by $ \pi_{\theta}(f_t, n_t) $, as
\begin{equation}\label{eq:encoder}
	A_t^s, A_t^s = \pi_\text{MPC}(E_S, E_N, E_V).
\end{equation}

Navigation embedding $E_{t}$ is used as a switch to activate various multilayer perceptron branches for inferring $A_{t}$ in CIL \cite{Codevilla:2018} and LBC~\cite{Codevilla:2019}. 
However, in our approach, we use a transformer to process the positional embedding $ E_V $.
To simplify the learning process, we utilize the navigation command $n_{t}$ as additional inputs.
%
We leverage the self-attention method in transformer to effectively extract multi-view feature\cite{Vaswani:2017}.
This approach allows the model to extract the relationships among remote frame patches, enabling it to link representation for horizontal perspectives of vehicle.

In MPC, a traffic representation is learned utilizing the transformer module with multiheaded self-attention layer~\cite{Vaswani:2017}, normalization, and feed forward multilayer perceptron modules. 
The caudal prefrontal cortex (CPC) is crucial for searching, identifying, and locating specific targets, ensuring that we can accurately focus on and navigate towards these targets\cite{lawler1987role}. 
These regions work together in the decision-making process, forming a highly integrated latent feature vector that effectively captures the essential information needed for making decisions\cite{wardak2004deficit}. 
This latent vector is then processed through a carefully constructed hidden layer, which generates accurate and reliable outputs, such as driving actions, value function estimates, and speed control commands.
The final output is obtained by linearly projecting the merged features for a self-attention head, which are passed to the CPC module.
We utilize 4 layers, and a layer with 4 attention modules. 
The middle size in the transformer module is matched with the output size of the ResNet.
%
The output from the transformer encoder is average pooled and passed through a multilayer perceptron.
This multilayer perceptron includes 3 FC with nonlinearity between layers.
The driving hehavior $A_t$ represents the vehicle acceleration $A_t^s$ (throttle and brake) and the wheel angle $A_t^s$ as depicted in Equ.~\ref{eq:encoder}.

\subsection{Loss Function}

\hspace{1pc}The BID network is designed to replicate the advanced information processing abilities of the human brain by carefully aligning neural pathways with corresponding network modules. 
Through iterative updates, the network adjusts its parameters to better match human decision-making processes, ultimately improving its performance and accuracy.
%
Based on the inferred output $A_t$ and the expert's actual output $\hat{A}_t$, the model loss function is described as:
\begin{equation}\label{eq:loss}
	L(A_t, \hat{A}_t) 
	= \lambda_{s} \lVert A_{t}^s-\hat{A}_t^s \rVert_{1}
	+ \lambda_{a}\lVert A_{t}^a-\hat{A}_{t}^a\rVert_{1} \enspace ,
\end{equation}
where $\lambda_{s}$ and $\lambda_a$ represent the steering angle and vehicle acceleration loss weight.
$\lVert\cdot\rVert_{1}$ denotes the Manhattan distance.
The weights are set as $\lambda_{s} = \lambda_{a} = 0.5$. 
Both vehicle acceleration and wheel angle are constrained within the range of $[-1, 1]$.
A acceleration greater than zero denotes throttle and a acceleration less than zero indicates braking in these work.

In behavior cloning~\cite{Codevilla:2019}, they use velocity inference regularization incorporated into the loss function to address the inertial issue resulting from the high likelihood of the ego vehicle remaining stationary.
However, we did not encounter this issue, and therefore, the velocity inference module is not included in the proposed BID model.
It is sufficient to achieve a excellent driving result with a plain training loss depicted in Equ.~\ref{eq:loss}, even in unfamiliar environments.
Additionally, the outputs from both the brain-inspired perception network and the brain-inspired decision network are concatenated to form a latent feature that captures critical driving-related information. 
This latent feature is then passed through a hidden layer to map it to driving actions.

\section{Experiment}
%

\subsection{Datasets} \label{sec:Dataset}

\hspace{1pc}We conduct training and testing on Carla~\cite{Dosovitskiy:2017},
and assess the proposed BID model using existing standard datasets\cite{codevilla2019exploring}. 
Each dataset specifies its own training scenes and weather conditions for data collection, 
and tests the BID's performance in scenes never seen before. 
In behavior cloning dataset\cite{codevilla2019exploring}, they focuses on transfer ability from Town01, a small scene with numerous T junctions and a variety of buildings, to Town02, a variant of Town01 with a mixture of residential and commercial buildings.
The offical test suit presents a challenging transfer problem across 6 scenes with various scenarios, including stop signs, lane changes, US-style junctions and roundabouts.
Based on the behavior cloning dataset~\cite{codevilla2019exploring}, we test the transfer ability between different weather conditions, though only two training weather types are evaluated to save computational resources. 
The behavior cloning benchmark includes three actor amount levels for each scene from less to more. 
A crowded configuration in behavior cloning is utilized, which uses a suitable number of actors to prevent typical jam with a large number of actors. 
In the offical test suit, we adjust actor number to meet the requirement of configuration.

As with the newest state-of-art model, we use the driving expert driver referred to reinforcement learning imitation~\cite{Zhang:2021} using online dataset with simulator.
This end-to-end method with reinforced feature exhibits very real action compared to a well-designed model in traffic scene. 
It is important to notice that the hoiminine driving demonstration will be used as training data in real world tests.
The standard configuration in~\cite{Zhang:2021} is used, which means that it is similar to the student driver in RL-Coach\cite{Zhang:2021} and MBI~\cite{Hu:2022}, the driving agent is the model found in simulator.
The frame from all the 3 horizontally concatenated vehicle-based RGB image sensors is 900 wide and 300 high, with a field of view of $180^{\circ}$.
The cameras are arranged without overlap, collectively covering a total filed of field.

Using the experienced agent, vehicle-based cameras and hero vehicle, driving data is collected in progressively more crowed scenarios.
Firstly, data is collected in Carla's Town01, a small scene that allows contains one lane merely, meaning we can't change lanes.
Specifically, 1620K frames approximately at 10 fps from vehicle-based RGB sensor is utilized on 4 different learning weather types.
For generalization testing, we use Carla's Town02, evaluating the agent under SoftRainSunset and WetSunset weather conditions. 
Next, the benchmark is collected on several simulator scene to capture crowed driving scenes, such as T junctions, highway entrances, exits, and crossing the intersections. 
To maintain consistency with setup in MBI~\cite{Hu:2022}, Town05 is reserved for testing, and totaling approximately 2,100K image per sensor is gathered at 10 fps in Town01 to Town06. 

\subsection{Implementation}

\hspace{1pc}The AdamW optimizer with a step size of $2 \times 10^{-4}$ at each iteration and weight decay of $0.02$ is utilized to training the BID model. 
We reduce the learning rate by half at 25, 45, and 70 epochs.
The batch size is 100, 
and model training is performed for 100 epochs on a single NVIDIA RTX A6000 GPU.

In ventral stream, after the input is processed by V2$_\text{COR}$ once, the resulting output is inputted back to V2$_\text{COR}$.
This process forms a recurrent connection.
We run V2$_\text{COR}$ and IT$_\text{COR}$ two times, and execute V4$_\text{COR}$ four times repeatedly, respectively.
This configuration yielded a best model performance based on our evaluation scores.
Similar to ResNet-50, a convolutional operation is accompanied by layer normalization and a rectified linear unit.
Layer normalization is unique to each time step.
The existing description of the ventral pathway does not include the connection that span across different regions, and retinal and lateral geniculate nucleus processing are not explicitly modeled.

\begin{figure}[t]
	\centering
	\includegraphics[width=1.0\linewidth]{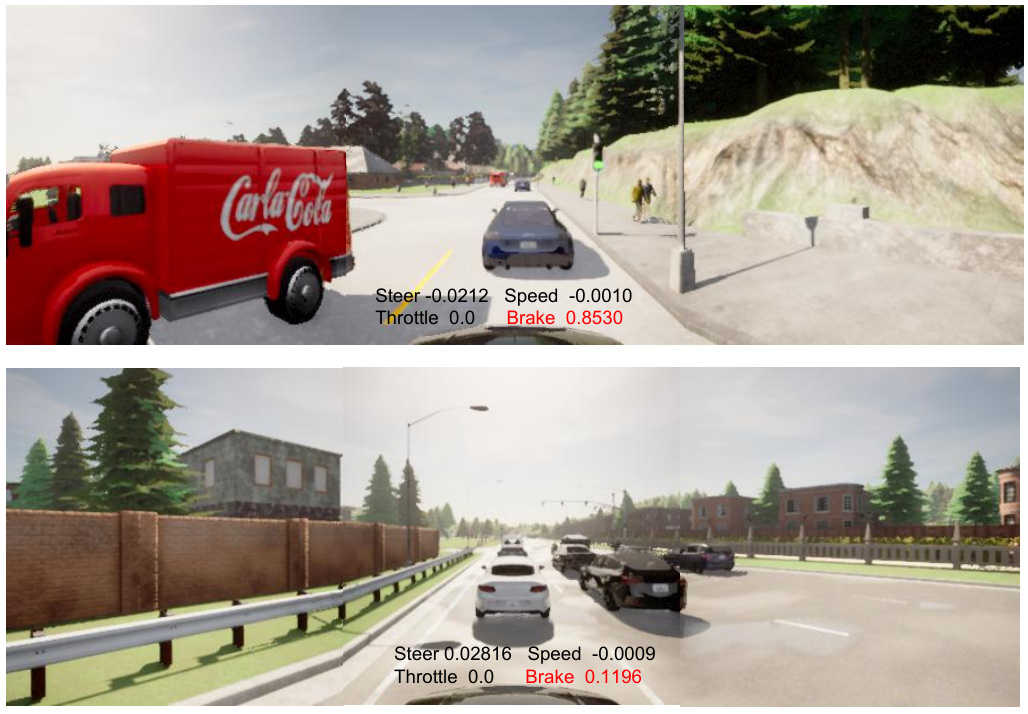}
	\caption{Top: When the ego-vehicle is driving on the road, even though the triffic light is green, the vehicle in front has stopped, but the ego-vehicle still brakes.
		Bottom: When the vehicle is moving and encounters a traffic jam in front, it will automaitcally brake and stop.
		The brake command in the figure is displayed in \textcolor{red}{red}.}
	\label{fig:command_ambiguous}
\end{figure}

\begin{figure}[ht!]
	\centering
	\includegraphics[width=\linewidth]{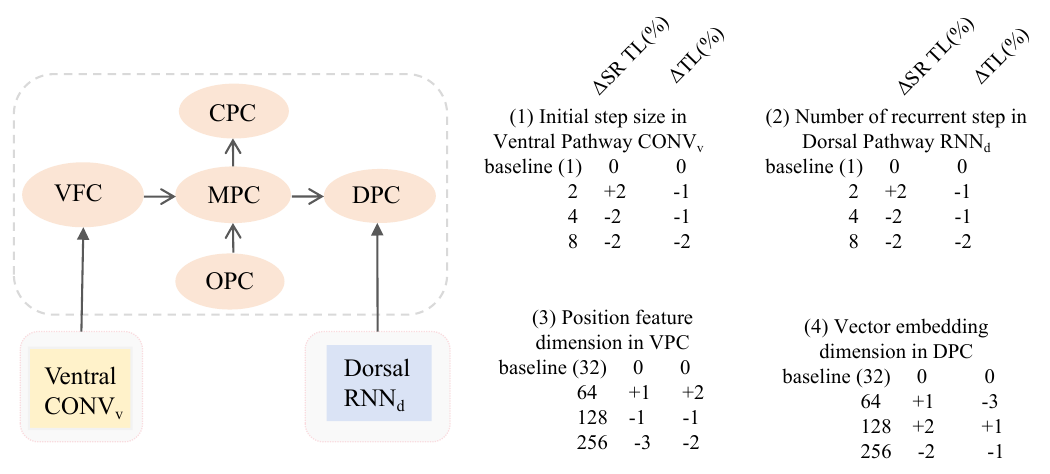}
	\caption{BID architecture analysis. 
		We analyse four main factors including the initial step size in ventral pathway, the number of recurrent step in dorsal pathway, the position feature dimension in VPC and the vector embedding dimension in DPC. 
		Each row presents how \emph{SR} and \emph{TL} score change based on BID when the hyperparameter changes. }
	\label{fig:structure_analysis}
\end{figure}

\subsection{Driving Evaluation}
\label{sec:Metrics}
\hspace{1pc}The BID agent's capabilities are constrained by the proficiency of the expert it is emulating. 
The success of the agent hinges on the expertise and skill level of its human counterpart.
If the expert performs poorly, the agent that mimics it will also show suboptimal results. 
The BID model is designed to closely follow the visual information processing system of the human brain, both structurally and functionally. 
When optimized, the model's output closely matches the expert's performance, as evaluated by similarity metrics.

We conduct experiments on small single-lane town following the nocrash dataset~\cite{Zhang:2021,Hu:2022}, and on multiple towns (Sec.~\ref{sec:small_town_results}) based on the offline Carla leaderboard dataset~\cite{Zhang:2021,Hu:2022}.

\begin{table*}
	\caption{Results in Town02.
		The expert refers to reinforcement learning~\cite{Zhang:2021}, while RL-Coach refers to Coach imitation learning. 
		Carla is used to evaluate all methods.
		3 experiment with various random seeds are used to calculate the mean values and standard deviations.
		Better performance is indicated by higher values for metrics marked with $\uparrow$, while lower are perferred for those marked with $\downarrow$.}
	\centering
	\resizebox{0.95\linewidth}{!}{
		\begin{tabular}{@{}lccc|ccc|ccc@{}}
			\hline
			& \multicolumn{3}{c}{None} & \multicolumn{3}{c}{Normal} & \multicolumn{3}{c}{Crowded}  \\
			
			& SR(\%,$\uparrow$) & SSR(\%,$\uparrow$) &  TL($\downarrow$)  & SR(\%,$\uparrow$) & SSR(\%,$\uparrow$) &  TL($\downarrow$) & SR(\%,$\uparrow$) & SSR(\%,$\uparrow$) & CV($\downarrow$) \\
			\hline
			
			RL-Coach\cite{Zhang:2021}  & $100\pm0.0$ & $85\pm1.2$ & $66\pm5.0$ & $97\pm2.3$ & $86\pm7.2$ & $66\pm54$ & $81\pm5.0$& $68\pm7.2$ & $63\pm52.7$\\
			MVA\cite{xiao2023scaling} & $100\pm0.0$ & $100\pm0.0$ & $0\pm0.0$  & $99\pm2.3$ & $97\pm3.1$ & $7\pm7.9$ & $83\pm7.6$ & $77\pm7.6$ & $45\pm21.5$ \\
			Our BID & $100\pm0.0$ & $100\pm0.0$ & $0\pm0.0$  & $99\pm2.3$ & $97\pm3.1$ & $7\pm7.9$ & $83\pm7.6$ & $77\pm7.6$ & $45\pm21.5$ \\
			\hline
			Expert & $100\pm0.0$ & $100\pm0.0$ & $0\pm0.0$ & $100\pm0.0$ & $97\pm0.0$ & $13\pm4.6$ & $84\pm2.0$ & $82\pm2.0$ & $37\pm14.1$ \\
			\hline 
		\end{tabular}
	}
	\label{tab:T2_NC_results}
\end{table*}

\begin{table*}
	\caption{Town05 outcomes based on offline measurements.
		Carla is used to evaluate all methods.
		3 experiment with various random seeds are used to calculate the mean values and standard deviations.
		Better performance is indicated by higher values for metrics marked with $\uparrow$, while lower are perferred for those marked with $\downarrow$.}
	\centering
	\resizebox{0.87\linewidth}{!}{
		\begin{tabular}{@{}lccccccccccccccccccccc@{}}
			\hline
			& $\uparrow$ APA(\%) & $\uparrow$ ADG
			& $\downarrow$ CV & $\downarrow$ CL & $\downarrow$ TL & $\downarrow$ OL & $\downarrow$ RD 
			\\
			\hline
			RL-Coach\cite{Zhang:2021}  & $92\pm3.1$ & $51\pm7.9$ 
			& $7.5\pm1.3$ & $4.3\pm1.6$ & $26.0\pm8.9$ & $5.4\pm2.7$ & $3.0\pm3.2$ & \\
			MBI~\cite{Hu:2022}
			& $98\pm2.2$ & $73\pm2.9$ 
			& $6.0\pm3.7$ & $0.0\pm0.0$ & $3.6\pm3.8$ & $3.5\pm1.5$ & $0.0\pm0.0$ \\   
			MVA\cite{xiao2023scaling} 
			& $98\pm1.7$ & $68\pm2.7$ 
			& $6.0\pm0.5$ & $3.8\pm0.7$ & $5.8\pm5.1$ & $6.1\pm2.2$ & $9.4\pm3.6$ \\
			Our BID 
			& $98\pm1.7$ & $68\pm2.7$ 
			& $6.0\pm0.5$ & $3.8\pm0.7$ & $5.8\pm5.1$ & $6.1\pm2.2$ & $9.4\pm3.6$ \\
			\hline
			Expert 
			& $99\pm0.8$ & $89\pm1.7$ 
			& $3.2\pm1.1$ & $0.0\pm0.0$ & $1.3\pm0.4$ & $0.0\pm0.0$ & $0.0\pm0.0$ \\
			\hline
		\end{tabular}
	}
	\label{tab:T5_results}
\end{table*}

\subsubsection{Official Measurement}\label{lb_metrics}
\hspace{1pc}We utilize the officiial measurement in several simulation scenes in order to match the evaluation with Carla leaderboard\cite{Hu:2022}.
The average path accomplishment (\emph{APA}) and the average drive grade (\emph{ADG}) are the most important measurement index. 
While \emph{APA} measures the average distance that the agent can run toward the goal,
\emph{ADG} penalizes driving performance according to the criteria outlines for the Carla measurement.

\subsubsection{Goal Instruction} 
\hspace{1pc}During training, we employ basic goal instructions like ``turn left" or ``go straight" when reaching a crossroad, just like in LBC\cite{Codevilla:2019}.
After passing a crossroad in more complex scenes, the ego-vehicle maybe legal when entering one of the several available routes. 
Because these information is avaiable through the global navigation system, if the vehicle agent departs from the pre-designed trajectory by entering a different lane, a corrected command is given, such as ``turn right" or ``turn left" as quickly as feasible. 
The corrected method is applied during evaluating merely.
An example of this process is shown in Fig.~\ref{fig:command_ambiguous}.

\subsubsection{Nocrash Benchmark Evaluation}\label{nocrash_metrics}

\hspace{1pc}Based on the quantity of variational actors ({\ie}, human, cars) in Carla scene, the benchmark comprises 3 mission with varying degrees of challenge: crowded, normal and none.
Town02 specifies that there are zero human and zeros cars (none); fifty human and fifteen cars (normal); and one hundred fifty human and seventy cars (crowded).
%
The initial actors quantity in nocrash benchmark frequently causes crowded and congestion at intersections in the crowded scenario~\cite{Zhang:2021}.
To address this, we adopt the \emph{crowded} scenarios described in RL-Coach~\cite{Zhang:2021}, reducing human population from one hundred and fifty to seventy. 

The success rate (\emph{SR}), which represents the proportion of runs that are normally finished, is the primary evaluation index for comparing agents.
We also offer the rigorous \emph{SR} (\emph{SSR}) for a more detailed measurement, which calculates the proportion of normal runs under a 0 violation policy for any rules, like running at a no passing traffic sign or deviating from planned route.
Additionally, we include other violative measurement.
The quantity of vehicle that fails to stop at a red traffic sign is represented by \emph{TL}.
The quantity of crashes with other cars is denoted by \emph{CV}.
The quantity of path inconformity where the high level navigation is not properly run is denoted by \emph{RD}.
\emph{OL} is the quantity of the agent deviate from its lane ({\eg}, to the sidewalk or to the opposite lane).
The quantity of crashes with scene is symbolized by \emph{CL}.

\subsection{Experiment Result}
\label{sec:Results}
\hspace{1pc}We contrast BID with three state-of-the-art, vision-based AD models: the Coach IL method (here RL-Coach) \cite{Zhang:2021}, MBI~\cite{Hu:2022} and MVA\cite{xiao2023scaling}. 
It is important to note that while BID employ sample produced by the RL-Coach method, optimizing a BID doesn't need image annotated by humans. 
In this instance, the RL-Coach model acts as the experienced driver during sample collection.
On the other hand, BID is trained using segmented Bird's Eye View (BEV) data as input, whereas BID model needs supervised signal from RL-Coach driver, who learned from real world data, while MBI is optimized.

\subsubsection{One lane Scenes} \label{sec:small_town_results}

\begin{figure*}[t]
	\centering
	\includegraphics[width=0.99\textwidth]{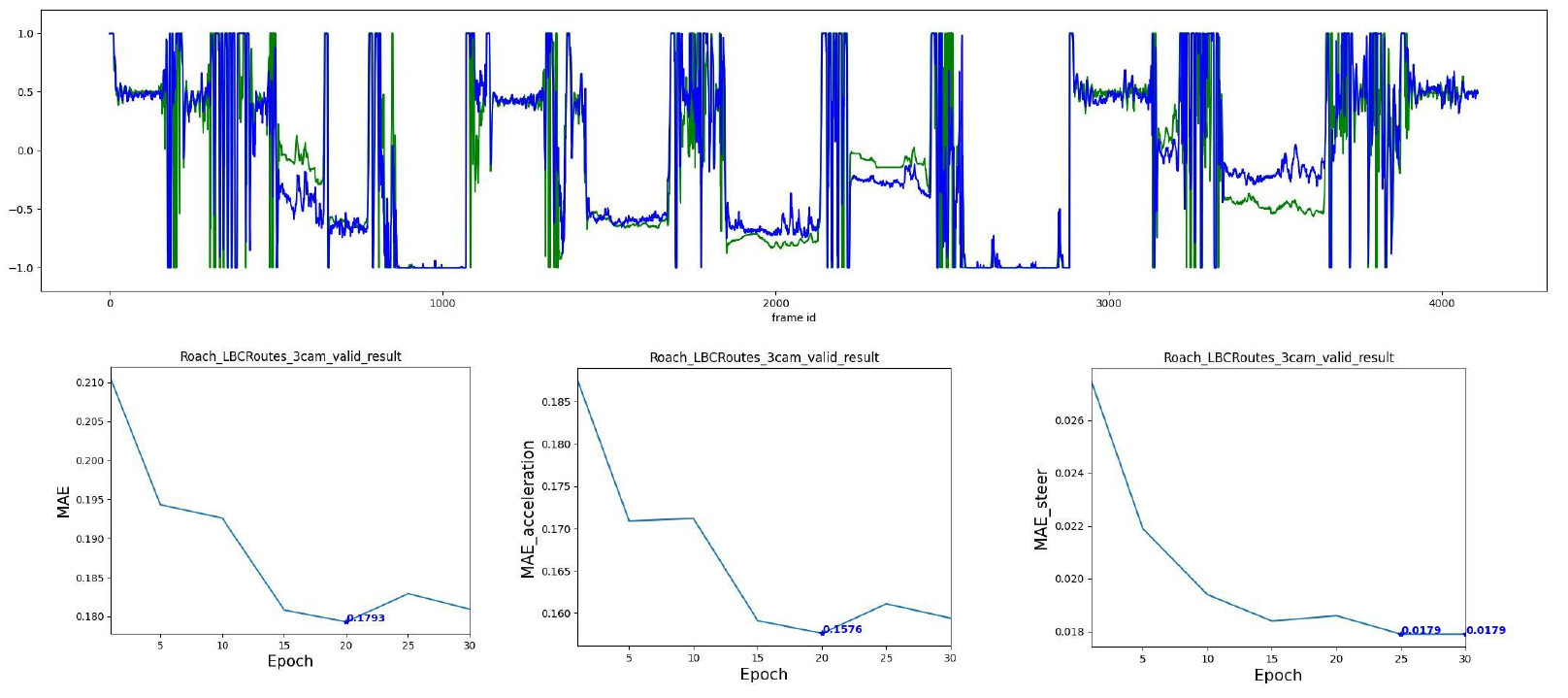}
	\vspace{-1ex}
	\caption{\textbf{Driving performance and training progress of BID.} 
		All BID agents are evaluated in LBCRoutes after 30 peoch.
		Top figure: The ground truth and model prediction for steering angles are shown in green and blue, respectively.
		Bottom figure: The errors in training progress are displayed every 5 epochs, including the mean absolute error (MAE) for steering and acceleration.
		While other metrics are tested with single random number, we describe the outcome as the averaged value over 5 test random number.
		For evaluation, the offical benchmark is employed.}
	\vspace{-1.5ex}
	\label{fig:score_eu_lb_tt_tn}
\end{figure*}

\hspace{1pc}Using one lane scenes in Carla and  the nocrash evaluations, firstly we perform preliminary experiments (Sec.~\ref{nocrash_metrics}). 
We employ Town01 to train BID, and use Town02 to evaluate our proposed model(Sec.~\ref{sec:Dataset}). 
MBI offers one that has been optimized on multiple simulator scenes, while no model is available that is specifically trained on Town01 alone. 
On the other hand, RL-Coach is a model optimized with various simulator scenes. 
We employ RL-Coach's model optimized on Town01 merely to ensure a reasonable evaluation. 
The \emph{SR} and \emph{SSR} with the different actor quantities (none, normal, crowded) are shown in Table~\ref{tab:T2_NC_results}. 
To provide a targeted assessment, we report \emph{IRL} (infractions with red light) in the none and normal scenarios, and \emph{CWV} (collisions with vehicles) merely for the crowded case in Table~\ref{fig:score_eu_lb_tt_tn}. 
It is important to emphasize that scenarios with fewer or no dynamic obstacles are more effective for evaluating the agent's response to red traffic signs, whereas crashes are appropriately assessed in towns with higher traffic density.

Generally speaking, BID performs the best performances across various missions. 
When it comes to avoiding traffic sign violations in the none case, BID performs noticeably better than RL-Coach, which raises the \emph{SSR}.
This conclusion holds true in a normal scenario as well. 
BID comes close to matching the expert's performance in the crowded scenario, outperforming RL-Coach in terms of \emph{SSR} and resulting in fewer crashes with cars again. 
According to the ground truth, scenario that failed in crowded runs are primarily cause of actors blocking up, which cause a timeout during path finish. 
Despite this, the expert's performance can still be considered a valid upper bound.

\subsubsection{Multi-scene Transferability}\label{sec:multi_towns_result}
\hspace{1pc}We evaluate the result of BID in increasingly intricate runs that Carla numerous scenes provide in the part. 
We use Carla official benchmark measurements (Sec.~\ref{lb_metrics}) and align the train and test settings with those used in MBI~\cite{Hu:2022}, as outlined in Sec.~\ref{sec:Dataset}. 
Table~\ref{tab:T5_results} displays the outcomes for every method that was optimized using multi-scene dataset. 
Out of the 3 methods, RL-Coach performs the least, incurring more violations and resulting in a noticeably smaller \emph{ADG}. 
APA of 98\% attained by BID is comparable to MBI. 
However, MBI has the highest \emph{ADG} score (73\%), while BID achieves 68\%. 
We attribute this difference to the fact that MBI rarely drives outside the pre-designed lane, as it is provided with the route map as input. 
In contrast, BID does not use an explicit route map and instead relies solely on high level  goal instructions.

\begin{table}
	\caption{\textbf{Camera-based imitation learning agent's \emph{SR} on crowded environments.}
		The mean value and deviation are computed based on 5 random seeds. 
		NCC: nocrash-crowded. ts: train scene. tw: train scene and weather have never seen before. ns: new scene. nw: new scene and weather have never seen before.
		In EDA, ``(E)" denotes an assembly of iteration sets, while ``+" indicates the additive jitter on the benchmark.}
	\setlength{\tabcolsep}{6.67pt}
	\centering
	\begin{tabular}{lccccc}
		\hline
		\emph{SR} \% $\uparrow$
		& NCC-tw  & NCC-ts   & NCC-nw  & NCC-ns  \\ 
		\hline
		SMN \cite{zhao2021sam} & 
		$48 \pm 4$ & $53 \pm 2$  & $30 \pm 3$ & $30 \pm 2$  \\
		LC \cite{chen2020learning} & 
		$64 \pm 2$ & $69 \pm 4$  & $40 \pm 5$ & $52 \pm 2$  \\
		LSD \cite{ohn2020learning} & 
		N/A & N/A & $31 \pm 4$ & $31 \pm 3$  \\
		EDA \cite{prakash2020exploring} & 
		$57 \pm 2$ & $65 \pm 4$  & $36 \pm 1$ & $37 \pm 2$  \\
		EDA\textsuperscript{+} \cite{prakash2020exploring}  & 
		$61 \pm 2$ & $61 \pm 2$  & $26 \pm 2$ & $35 \pm 1$  \\
		Baseline, $\mathcal{L}$ & 
		$30 \pm 2$ & $\mathbf{87} \pm 2$  & $29 \pm 5$ & $35 \pm 9$  \\
		Improved, $\mathcal{L}_\text{D} $ & 
		$31 \pm 1$ & $\mathbf{87} \pm 1$  & $29 \pm 4$ & $33 \pm 10$  \\
		Proposed, $\mathcal{L}_\text{D}+\mathcal{L}_\text{N}$ & 
		$\mathbf{83} \pm 3$ & $86 \pm 3$  & $\mathbf{79} \pm 1$ & $\mathbf{79} \pm 4$  \\
		\hline
	\end{tabular}
	\vspace{-1ex}
	\label{table:sucess_rate_nc_dense}
	\vspace{-2ex}
\end{table}

\begin{table*}
	\caption{\textbf{Analyzing BID's output effect and infractions in scene never seen it before with a lot of people.} 
		IP: infraction penalty. CWO: collision with others. CWP: collision with pedestrian. 
		CWV: collision with vehile.
		IRL: infraction with red light.
		ABD: agent blocked.
		5 test random number seeds are employed to display the mean value and deviation.}
	\setlength{\tabcolsep}{7.4pt}
	\centering
	\begin{tabular}{lccccccccc} 
		\hline
		& \begin{tabular}{@{}c@{}} DS ($\uparrow$) \end{tabular} 
		& \begin{tabular}{@{}c@{}} SR ($\uparrow$)  \end{tabular} 
		& \begin{tabular}{@{}c@{}} IP ($\uparrow$) \end{tabular} 
		& \begin{tabular}{@{}c@{}} PA ($\uparrow$)  \end{tabular} 
		& \begin{tabular}{@{}c@{}} ABD ($\downarrow$) \end{tabular} 
		& \begin{tabular}{@{}c@{}} IRL ($\downarrow$) \end{tabular} 
		& \begin{tabular}{@{}c@{}} CWO ($\downarrow$) \end{tabular}  
		& \begin{tabular}{@{}c@{}} CWP ($\downarrow$) \end{tabular}  
		& \begin{tabular}{@{}c@{}} CWV  ($\downarrow$) \end{tabular}  \\
		\hline
		$\mathcal{L}_\mathrm{b}$
		& $42 \pm 3$ & $32 \pm 5$  & $76 \pm 4$ & $61 \pm 5$  
		& $19.4\pm 14.4$ & $3.33 \pm 0.58$  & $0.53 \pm 0.55$  & $\mathbf{0}\pm0$ & $0.63 \pm 0.50$    \\
		$\mathcal{L}_D$
		& $65\pm3$ & $58\pm6$  & $76\pm1$ & $85\pm2$  
		& $2.83\pm1.46$ & $1.5\pm0.2$  & $2.06\pm1.28$  & $\mathbf{0}\pm0$ & $1.37\pm1.10$    \\
		$\mathcal{L}_N$
		& $90\pm2$ & $85\pm1$ & $75\pm2$  & $78\pm0$  
		& $3.36\pm0.21$ & $0.69\pm0.06$ & $0.51\pm0.25$  & $\mathbf{0}\pm0$  & $0.52\pm0.17$    \\
		$\mathcal{L}_D+\mathcal{L}_N$
		& $89 \pm 3$ & $88 \pm 5$  & $90 \pm 3$ & $\mathbf{97} \pm 0$  
		& $0.83 \pm 0.03$ & $0.62 \pm 0.22$ & $\mathbf{0.07} \pm 0.04$  &  $0.01 \pm 0.01$ & $0.22 \pm 0.07$    \\
		$\mathcal{L}_{BID}$
		& $\mathbf{96} \pm 2$ & $\mathbf{96} \pm 3$ & $96 \pm 0$ & $97 \pm 2$ 
		&  $\mathbf{0} \pm 0$ & $\mathbf{0.14} \pm 0.18$  & $0.12 \pm 0.08$  & $\mathbf{0} \pm 0$  & $\mathbf{0.03} \pm 0.06$  \\
		Expert
		& $91 \pm 1$ & $78 \pm 2$ & $\mathbf{97} \pm 1$ & $81 \pm 2$ 
		& $0.19 \pm 0.07$ & $1.92 \pm 0.22$  & $0.19 \pm 0.07$ & $\mathbf{0} \pm 0$ & $0.17 \pm 0.09$   \\
		\hline
	\end{tabular}
	\vspace{-1ex}
	\vspace{-2.5ex}
	\label{table:infraction}
\end{table*}

\subsubsection{BID's Activation Visualization}
\label{sec:Visualization}
\hspace{1pc}We want to know which parts of the image BID focuses on during decision-making. 
To achieve this, we use Grad-CAM~\cite{Selvaraju:2017}, where gradients from driving behavior embedding to last convolution in ResNet-50. 
The process generates the heatmap that identifies the areas of frame that are most crucial for driving behavior generation. 
Nevertheless, Grad-CAM was initially created in frame recognition task, where outcomes are often positive, whereas BID handles regressive mission, so the outcome can take both positive and negative values.
We can't just concentrate on the representation map's positive gradients to modify Grad-CAM for these situation.
Instead, the calculation is split into 2 scenarios according to the outcome value's symbol. 
We employ negative gradient to compute a feature map weights when the steering angle or acceleration is negative, and positive gradients are used when the output is positive.

\begin{figure}[ht!]
	\centering
	\includegraphics[width=\linewidth]{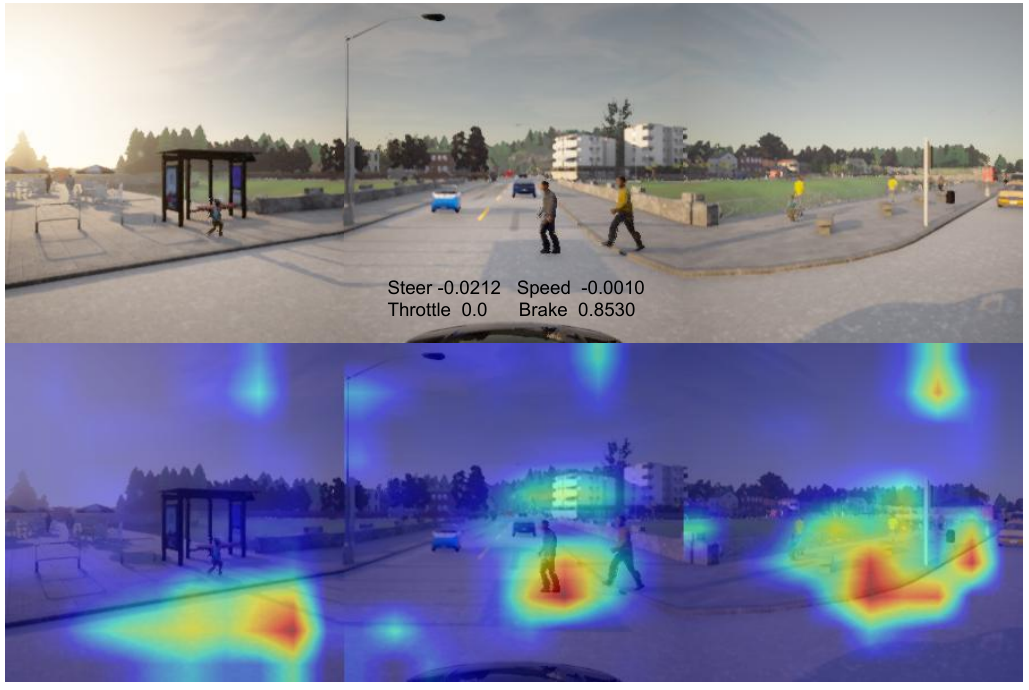}
	\caption{Activation maps of BID at an intersection in Town02 show three highly activated image areas from driving viewpoints: 
		the lane shoulder in the right, the crossing pedestrians in the center, and the driving area in the left. 
		The causality between observation and action is demonstrated by a strong braking response (0.8530) due to the presence of pedestrians, despite the ``go-straight" command. 
		This highlights BID's ability to prioritize pedestrian safety over other factors in BID's decision-making process. }
	\label{fig:attention_ped_greed}
\end{figure}

Fig.~\ref{fig:attention_ped_greed} illustrates the attention image in a crossroad. 
3 regions of the frame are most noticed: a lane shoulder on the frame right, the human crossing the street in the frame center, and the driving area on the frame left. 
This suggests that BID demonstrates a thorough comprehension of the scenario and a discernible connection between its observations and actions. 
Specifically, BID chooses to slow down because of the crossing human, despite the green traffic sign and the goal instruction is to turn left. 
This indicates that BID appropriately prioritizes the presence of pedestrians over the traffic signal and navigation command.

\subsubsection{Ablation}

\hspace{1pc}An imitation learning ego-vehicle's ability to perform is inherently constrained by the quality of experienced driver it imitates. 
Comparing imitation learning ego-vehicle trained on an experienced driver is pointless if that driver performs poorly.
This issue becomes apparent in a scenario never seen it before with crowded actors, where the performance of autopilot typically is poor. 
We optimize the BID (Fig.~\ref{fig:score_eu_lb_tt_tn}) and conduct ablation studies with architecture analysis (Fig.~\ref{fig:structure_analysis} and the crowded actors configuration (Table~\ref{table:infraction}) to establish a high precision effect and ensure a effective evaluation, allowing the proposed agent to attain a \emph{DS} of 96\%. 
As shown in Table~\ref{table:sucess_rate_nc_dense}, the optimal configuration from the ablative experiments is also assessed on benchmark in crowded actors for a more meaningful comparison with latest models.

Fig.~\ref{fig:score_eu_lb_tt_tn} shows the behavior grades of experienced and imitation learning drivers for an epoch on offical benchmark with crowded actors.
The basic $\mathcal{L}$ represents the proposed implementation of the baseline BID trained by expert. 
According to the leaderboard instructions, this additional navigation vector helps separating scenes where the map's complexity may make it difficult to understand the labels of each frame.
Given the improvements in our best-performing BID, it is expected that $\mathcal{L}_D$ and $\mathcal{L}_D + \mathcal{L}_N$ gain better \emph{SR} than results described in Table~\ref{table:sucess_rate_nc_dense}.
The significant performance gap between the baseline and $\mathcal{L}_D + \mathcal{L}_N$, especially when transferring to the scene never seen it before, highlights the basic approach limitations.

$\mathcal{L}_D$ outperforms $\mathcal{L}_b$ overall by incorporating the dorsal stream and speed embedding into the baseline $\mathcal{L}_b$.
Additionally, learning from the action distribution allows $\mathcal{L}_D$ to generalize superior than $\mathcal{L}_b$ on the benchmark dataset, but not on the offical leaderboard.
When $E_S$ is given merely, the feature required to generate accurate driving action does representation mapping to enhance performance.
Since the expected path is given throught nagivation command in this instance, $E_N$ includes goal instructions.
The use of $E_N$ improves accuray for the benchmark because it encodes goal instruction, including the feature pointing to the following expected path.
To evaluate this idea, we utilize a integrated model structure in which the goal instruction described as one hot encoding is added to the evaluation feature $E_N$.
In the benchmark scene transferability evaluation, $\mathcal{L}_\text{BID}$ achieves the highest behavior degree in imitation learning ego-vehicles.

Adding supervised representation alongside feature matching accelerates the convergence of the process, as demonstrated by $\mathcal{L}_D + \mathcal{L}_N$.
However, when feature matching is omitted, supervised representation doesn't lead to better result with $\mathcal{L}_D$.
These suggests the possibility of value estimation and representation mapping working in tandem.
Predicting the value should be made easier by imitating the representation of RL-coach, which intuitively contains the feature needed for value prediction. 
In turn, these estimation can be enhanced by regularizing representation mapping, improving overall performance.

\subsubsection{Infraction and Result Analysis}

\hspace{1pc}Table~\ref{table:infraction} presents a particular result and infraction analyse on the benchmark dataset with crowded actors in the scene never seen it before. 
Interestingly, the remarkably better \emph{ABD} degree in the basic $\mathcal{L}_b)$ is primarily caused by results from heavy rains. 
The issue is significantly reduced by imitating expert leading to a 23\% absolute improvement in the \emph{DS} for $\mathcal{L}_D$.
While maintaining the same IL strategy, this gain illustrates the advantages of using a more knowledgeable driver. 
Better improvements are achieved with the addition of soft targets and latent feature supervision in $\mathcal{L}_D+\mathcal{L}_N$, resulting in another 30\% absolute increase in performance. 
By handling red lights more effectively, this BID agent achieves a driving score of 89\%, reaching expert-level performance with just camera image as input.

\section{Conclusion}
\label{sec:conclusion}
\hspace{1pc}This paper explores the application of the BID model in brain-inspired autonomous driving. 
The model is designed to replicate the way information is processed along the brain's visual and decision-making pathways, with the goal of improving the interpretability and robustness of autonomous driving systems by mimicking neural processes. 
In addition to visual cue perception, the BID model emphasizes optimizing decision-making strategies by simulating the decision-making process of the prefrontal cortex. 
This approach of mimicking human decision-making helps autonomous driving systems strike a balance between interpretability and precision, particularly in complex traffic scenarios, ultimately enhancing safety and system robustness.


\bibliography{BID.bib}
\bibliographystyle{IEEEtran}

\newpage
%

\vfill

\end{document}